\documentclass[letterpaper, 10 pt, conference]{ieeeconf}  

\IEEEoverridecommandlockouts                              

\title{\LARGE \bf
Grasping by Parallel Shape Matching
}

\author{Wenzheng Zhang$^{1,*}$, Fahira Afzal Maken$^{2}$, Tin Lai$^{1}$, Fabio Ramos$^{1,3}$
\thanks{$^*$ Corresponding author: wzha2981@uni.sydney.edu.au}
	\thanks{$^1$ School of Computer Science, The University of Sydney, Australia}
        \thanks{$^2$ Data61, CSIRO, Australia}
	\thanks{$^3$ NVIDIA, USA}
}

\usepackage[short]{optidef}
\usepackage{amsmath}
\usepackage{amssymb}
\let\labelindent\relax
\usepackage{enumitem}
\RequirePackage[activate={true,nocompatibility},final,tracking=true,kerning=true,spacing=true,factor=1100,stretch=10,shrink=10]{microtype}
\microtypecontext{spacing=nonfrench}\linepenalty=1000
\usepackage{mathptmx}
\usepackage{graphicx}
\usepackage{caption}
\graphicspath{{Images/}}
\usepackage[ruled, shortend, linesnumbered]{algorithm2e}

\usepackage{bm}
\usepackage{amssymb}
\usepackage{multirow}
\usepackage{siunitx}
\usepackage{booktabs}
\usepackage[font=small]{caption}
\usepackage{lmodern}
\usepackage{comment}
\usepackage[pagebackref=false,breaklinks=true,colorlinks,bookmarks=false]{hyperref}
\newlength{\textfloatsepsave} 

\usepackage[left=0.75in, right=0.75in, top=0.81in, bottom=0.65in]{geometry}
\begin{document}

\maketitle
\thispagestyle{empty}
\pagestyle{empty}

\begin{abstract}
\small
Grasping is essential in robotic manipulation, yet challenging due to object and gripper diversity and real-world complexities. Traditional analytic approaches often have long optimization times, while data-driven methods struggle with unseen objects. This paper formulates the problem as a rigid shape matching between gripper and object, which optimizes with Annealed Stein Iterative Closest Point (AS-ICP) and leverages GPU-based parallelization. By incorporating the gripper's tool centre point and the object's centre of mass into the cost function and using a signed distance field of the gripper for collision checking, our method achieves robust grasps with low computational time. Experiments with the Kinova KG3 gripper show an 87.3\% success rate and 0.926s computation time across various objects and settings, highlighting its potential for real-world applications.
\end{abstract}

\section{INTRODUCTION}
Grasping is a fundamental skill in robotic manipulation, essential for various applications, including everyday tasks such as cleaning. Despite extensive research, effective grasp generation remains challenging due to various objects, diverse gripper types, and the complexities of real-world environments.


There are two primary approaches for grasp generation \cite{c1}: analytic and data-driven methods. Analytic approaches use gripper and object models to determine optimal grasp positions, validated through simulations. However, discrepancies between models and real environments often lead to failures when transitioning to physical robots. Additionally, optimizing grasp configurations in high-dimensional spaces for multi-fingered hands is time-consuming, making it impractical for real-time usage. 

Data-driven approaches have gained significant attention due to their higher success rates and reduced pose generation times. One notable application is predicting the success of grasp poses using extensive datasets of successful grasps, which are either manually collected or generated through simulations. Despite their advantages, these approaches face challenges in generalizing to unseen objects that were not included in the training data.

Each approach mimics aspects of human grasping: data-driven methods use prior knowledge to generate potential grasp preshapes, while analytic methods adjust hand poses during grasping.

\begin{figure}[t]
    \centering
    \small
    \includegraphics[width=5cm]{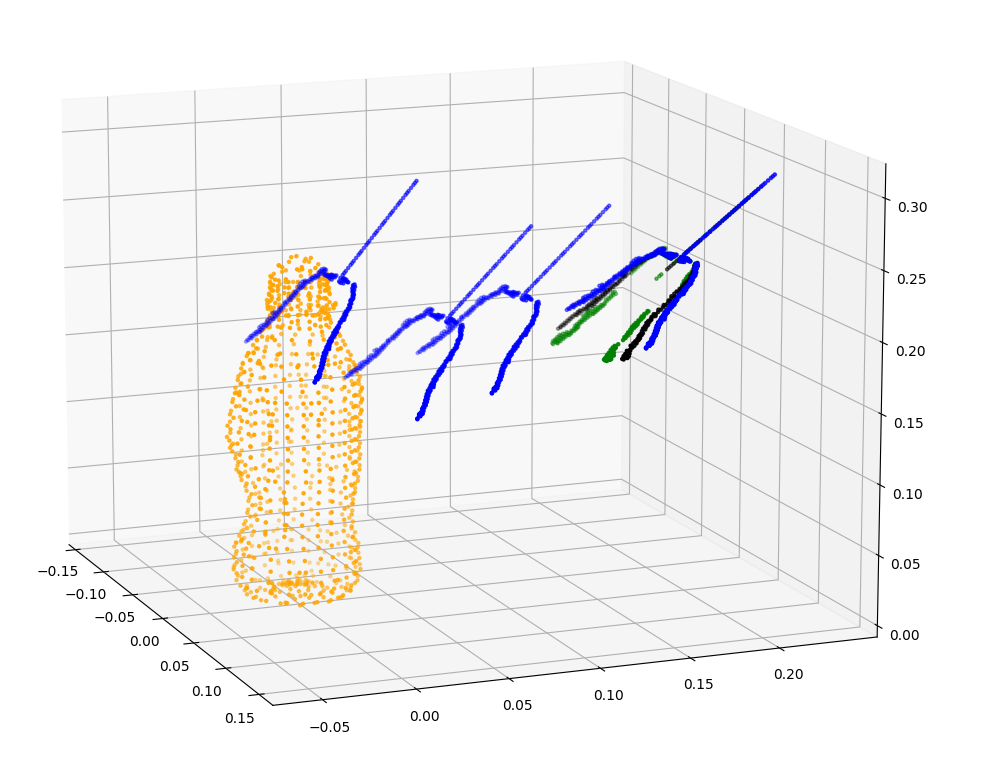}
    \setlength{\belowcaptionskip}{-15pt}
    \caption{An illustration of the optimization process. Green, blue and black point clouds are the initial poses of three preshapes of the KG3 gripper. Blue plots show the optimization process to find the grasp pose of a single preshape.}
    \label{fig:op_process}
\end{figure}

This paper addresses the grasping problem by formulating it as an optimization task where the final grasp pose is achieved by matching the point clouds of the gripper and the object. Our focus is on power grasps, where the grasping pose involves the entire palm of the gripper. We are not focusing on precision grasps, which involve only the fingers. Precision grasping requires careful selection of gripper preshapes, which we will leave for future work. The optimization steps are visualized in blue in  Fig~\ref{fig:op_process}. We initialized transformation parameters using the Fibonacci sequence \cite{c35} and parallelized the optimization process using the Annealed Stein Iterative Closest Point (AS-ICP) algorithm \cite{c24}\cite{c29} to ensure a better parameter distribution during the optimization process.

To achieve a stable grasp and minimize moments while holding the object, we incorporate the distance between the gripper's tool center point and the object's center of mass into the cost function. This ensures contact points are positioned to reduce rotational forces and increase stability. We also use a signed distance field of the gripper for collision checking during matching, preventing invalid grasp poses from non-reachable inner surfaces. Leveraging GPU parallel computing, our method significantly reduces optimization time and avoids local minima with multiple preshapes and initializations using AS-ICP.

The main contribution of this paper is a framework for generating grasp poses with rigid shape matching between the gripper and object's point clouds. We leverage GPU parallelization with optimization formulated as Annealed Stein ICP. The algorithm is independent of the gripper type and does not rely on costly finger joint angle optimization typical of analytic approaches. Unlike other ICP-based methods, we treat the gripper as a rigid body, using its signed distance field (SDF) for collision checking, which reduces dependency on high-precision object point clouds and normals. Our approach generates robust optimal grasp poses with uniformly distributed initializations, demonstrating robustness in simulations and complex real-world experiments with various objects and noisy, partial point clouds. For 11 objects using the Kinova KG3 gripper, we achieve an average success rate of 87.3\% 

\section{RELATED WORK}


The analytic approach relies on object and gripper models and selects the grasp pose based on various grasp quality metrics \cite{c4}. The Contact Point Optimization (CPO) method \cite{c5} combines Palm Pose Optimization (PPO) to generate grasp poses for a three-fingered hand based on a given pose of a parallel gripper. In \cite{c6}, this dual optimization technique is applied to generate grasp poses for objects in cluttered environments using a three-fingered hand. Iterative Surface Fitting (ISF)~\cite{c7} has been employed for grasping with customized parallel grippers. The algorithm is generalized to multi-fingered hands for both precision and power grasp in \cite{c8}. Additionally, the combination of multi-dimensional iterative surface fitting (MDISF) and grasp trajectory optimization (GTO)~\cite{c9} has been employed to generate optimal grasps using three-fingered hands. Optimizing in high-dimensional spaces for multi-fingered hands can be a time-intensive process. In contrast, our method simplifies this by bypassing the need to optimize finger joint angles, significantly reducing the time required to find a suitable grasp pose.

The data-driven approach relies on large labelled datasets and requires significant training time. However, it achieves higher success rates, particularly with objects similar to those used during the training. \cite{c11} utilizes a transformer model and achieves a success rate of 97.99\% on the Cornell dataset. Multilevel Convolutional Neural Networks have been proposed to grasp objects using multi-fingered hands, which achieves a success rate of 95.82\% for 16 objects \cite{c12}. \cite{c31} proposes AnyGrasp, which can generate grasp poses dynamically. \cite{c10} proposes AdaGrasp which learns a policy that can be used for different grippers. Most of these existing methods face challenges in generalizing to objects that are not present in the training datasets. In contrast, our approach eliminates the need for training and demonstrates an enhanced ability to generalize to unfamiliar objects. Moreover, our technique can be applied alongside these methods to enhance grasp quality.

Dexterity, equilibrium, stability, and dynamic behaviour are fundamental characteristics of grasp synthesis~\cite{c13}. These characteristics are formulated as grasp quality metrics, categorized into two main groups: contact points and hand configuration, further divided into 24 different quality indices \cite{c14}. To guide the selection and combination of metrics, a comprehensive evaluation of the variability, correlation, and sensitivity of ten selected metrics has been conducted \cite{c15}. A correlation study involving 16 metrics has been carried out for multi-fingered hands \cite{c16}.  Based on this study, the authors selected force closure, area of the grasp polygon, distance to the object's center of mass, and configuration of the finger joints as criteria for in-hand manipulation planning. In this paper, our approach utilizes a predefined preshape of the gripper as input, obtained by simultaneously closing fingers, as shown in Fig. \ref{fig:preshapes}. These preshapes are then optimized to minimize the matching error against the object. Our grasping cost function comprises the distance to the object's centre of mass and the matching error, which we describe in Section \ref{sec:method}.

\section{Preliminaries} 
In this section, we provide an overview of standard ICP~\cite{c17}, SGD-ICP~\cite{c2,c3} and Stein ICP~\cite{c24}, which constitute core components of our matching-based grasp method.

\subsection{Standard ICP}
Standard ICP~\cite{c17} aligns two point clouds by iteratively updating the initial transformation $\theta_0 = \{t, q\}$ between two point clouds. Here $t =\{x, y, z\}$ represents the 3D translation and $q =\{q_w, q_x, q_y, q_z\}$ represents the unit quaternion, with $q_w$ being the scalar and $\{q_x, q_y, q_z\}$ the vector part of the quaternion. The algorithm proceeds through the following two matching and minimization steps until convergence is achieved:
\begin{enumerate}[leftmargin=*]
  \item \emph{Matching Step:} pairs the transformed source point cloud, $\mathcal{S'} = \{s_i'\}_{i=1}^N$, with the reference point cloud $\mathcal{R} = \{r_i\}_{i=1}^M$ on the basis of a distance metric, where $s_i$ and $r_i \in  R^3$ are $N$ and $M$ points in 3D space.  The commonly used point-to-point distance metric finds the pair with the nearest neighbour as follows:
\begin{equation}
\hat{r}_i =  \operatorname*{argmin}_{r_j \in \mathcal{R}}\|s_i' \textendash r_j\|
    \label{eq:matching}
\end{equation}
where $\hat{r}_i$ is the closest point to $s_i' = (R s_i + t)$, $R \in \mathbb R^{3 \times 3}$ is the rotation matrix parametrized by ${\theta}^{4:7}$, and $t \in \mathbb R^{3 \times 1}$ is the translation vector comprising of ${\theta}^{1:3}$.
Henceforth, we use the notation ${\theta}^{i:j}$ to denote a sequence of theta values spanning from index $i$ to index $j$, inclusive.

\item \emph{Minimization Step:} updates $\theta_k$ to minimize a loss function defined by the distance between the paired points in the source and reference point clouds. 
The updated equation for the point-to-point distance metric is defined as follows: 
\begin{equation}
\theta_{k+1} =  \operatorname*{argmin}_{\theta}\frac{1}{N}\sum_{i}^{N}\|R_ks_i+t_k\textendash \hat{r}_i\|^2
\label{eq:minimization}
\end{equation}
where $k$ is the iteration number. Equation~\eqref{eq:minimization} can be solved in closed-form using either Horn's method \cite{c18} or singular value decomposition (SVD) \cite{c19}.
\end{enumerate}
\subsection{Stochastic Gradient Descent ICP}
SGD-ICP~\cite{c2,c3} solves the optimization problem of ICP ~\eqref{eq:minimization} with stochastic gradient descent (SGD) \cite{c20}. It samples a mini-batch $\mathcal{M}$ of $m$ points from the source cloud and computes the gradient of ~\eqref{eq:minimization} to update $\theta$ in the following equation:
\begin{equation}
    \theta_{k+1} = \theta_k \textendash \eta A\bar{g}(\theta_k,\mathcal{M}_k)
    \label{eq:sgdicp_update}
\end{equation}
where $\eta$ is the learning rate, $A\in R^{6\times6}$ acts as a pre-conditioner for the gradients, and $\bar{g}$ are the average gradients which are computed as follows:

\begin{equation}
\bar{g}(\theta_k^{1:3},\mathcal{M}_k) = \frac{1}{m}\sum_{i}^{m}((R_ks_i+t_k)\textendash \hat{r}_i)\frac{\partial t_k}{\partial \theta_k^{1:3}}
    \label{eq:trans_grad}
\end{equation}
\begin{equation}
\bar{g}(\theta_k^{4:7},\mathcal{M}_k) = \frac{1}{m}\sum_{i}^{m}((R_ks_i+t_k)\textendash \hat{r}_i)\frac{\partial R_k}{\partial \theta_k^{4:7}}s_i
    \label{eq:quat_grad}
\end{equation}
where $\bar{g}(\theta_k^{1:3},\mathcal{M}_k)$ and $\bar{g}(\theta_k^{4:7},\mathcal{M}_k)$ are the gradients of the cost function w.r.t translations and rotations respectively.

\subsection{Annealed Stein ICP}
To ensure a better distribution of $\theta$, we utilise Stein ICP developed in \cite{c24}, which uses SGD-ICP gradients in the Stein variational
gradient descent (SVGD) framework \cite{c25}. Stein ICP approximates an intractable but differentiable posterior distribution of transformation parameters $p(\mathbf{\theta})$ by constructing a non-parametric variational distribution represented by a set of $K$ particles $\{\mathbf{\theta}_j\}_{j=0}^K$. 
The SVGD algorithm~\cite{c25} provides the optimal update direction for particles: 
\begin{equation}
\phi^*(\cdot) = \mathbb{E}_{\theta\sim q} [\nabla_\theta \log p(\theta)k(\theta,\cdot) + \nabla_\theta k(\theta,\cdot)]
    \label{eq:svgd}
\end{equation}
In practice, the update rule for particles is given by:
\begin{equation}
\theta_j \leftarrow \theta_j + \eta \hat{\phi}^*(\theta_j).
    \label{eq:svgd-update}
\end{equation}
In Stein ICP, the gradients of the log of posterior $\nabla \log p(\theta)$ in~\eqref{eq:svgd} are replaced by the gradients of the log of the likelihood function ($\Bar{g}(\theta_j, \mathcal{M})$) from SGD-ICP(\eqref{eq:trans_grad},~\eqref{eq:quat_grad}) and the gradient of the log of priors ($\nabla_\theta \log p(\theta_j)$) as follow:
\begin{equation}
    \begin{aligned}
    \hat{\phi}^*(\theta) = \sum_{j=1}^{K} [{} &-\gamma(t)(N\Bar{g}(\theta_j, \mathcal{M}) + \nabla_\theta \log p(\theta_j))k(\theta_j,\theta)\\
&+ \nabla_\theta k(\theta_j,\theta)],
\label{eq:svgd-icp}
    \end{aligned}
\end{equation}
where $\gamma(t)$ is annealing schedule defined in equation ~\eqref{eq:annealing}.  $\nabla \log p(\mathbf{\theta})$ is represented by gradients from Gaussian priors for translations and von Mises priors for rotations.
The first gradient term in \eqref{eq:svgd-icp}, weighted by a kernel function, determines the steepest direction for the log probability. Conversely, the second term represents the gradient of the kernel function, acting as a repulsive force that promotes dispersion among particles and prevents them from converging to local modes of the log probability.
%
In this paper, we use RBF kernel for translation:
\begin{equation}
k(\theta_{1:3}', \theta_{1:3}) = \exp{(-\frac{1}{h}\|\theta_{1:3}-\theta_{1:3}'\|^2_2)},
    \label{eq:rbf}
\end{equation}

and the dot product as a kernel for rotation:
\begin{equation}
k(\theta_{4:7}', \theta_{4:7}) = \text{abs}(\theta_{4:7}\cdot\theta_{4:7}').
    \label{eq:dot}
\end{equation}

To address the mode collapse problem inherent in SVGD~\cite{c34}, we adopt annealed SVGD~\cite{c29} with SGD ICP. The annealing schedule in ~\eqref{eq:annealing} encourages more exploration, resulting in a method known as Annealed Stein ICP, which improves the distribution of poses crucial for our grasping tasks. 
\begin{equation}
\gamma(t) = (\frac{\mod(t, T/C)}{T/C})^p
    \label{eq:annealing}
\end{equation}
where t is the current iteration, T is the maximum iteration, C is the number of cycles, and p is an exponent determining the transition speed between the two phases. 
\section{Parallel Shape Matching}\label{sec:method}

\subsection{Problem Formulation}
We investigate grasp synthesis for unknown objects while avoiding the expensive optimization of finger joints.  Our algorithm takes as input a gripper's inner surface point cloud $\mathcal{S}$ (Fig. \ref{fig:preshapes} (b, d)) and the object's 3-D point cloud $\mathcal{R}$ for matching, a complete gripper's point cloud $\mathcal{G}$ (Fig. \ref{fig:preshapes} (a, c)) for generating SDF and object's point cloud with the table's point cloud $\mathcal{C}$ for collision checking. 
Specifically, we utilize point clouds of different finger configurations as preshapes, which remain unchanged during optimization. The algorithm aims to determine an optimal grasp pose parameterized by $\theta$. All transformations are with respect to the object's coordinate frame. The grasp pose is optimized using parallel AS-ICP with the following constrained loss: 
\begin{equation}
\begin{aligned}
&\min_{\theta_i} \quad \mathcal{L}(\mathcal{R},T_{\theta}(\mathcal{S})) \\
&s.t. \quad dist(\mathcal{C}, SDF(T_{\theta}(\mathcal{G}))) < 0,
 \label{eq:icp-objective}
\end{aligned}
\end{equation}
where $T_{\theta}(\mathcal{*})$ is the transformed point cloud w.r.t. $\theta$, and the loss function
\begin{equation}
   \mathcal{L} = \mathcal{L}_{com} + \mathcal{L}_{ct} 
    \label{eq:total_weighted_loss}
\end{equation}
is a combination of two grasp quality metrics, matching error ($\mathcal{L}_{ct}$) and distance between tool centre point and centre of mass ($\mathcal{L}_{com}$), which we will explain in the next Section~\ref{sec:cost}.

The loss function is constrained by the distance between the object's point cloud and the gripper's signed distance field (SDF) to ensure a physically plausible grasp pose. A collision occurs when the gripper's point cloud is inside the physically infeasible space of the object's point cloud, corresponding to a positive distance value from the SDF. 

\subsection{Cost Functions and Gradients}\label{sec:cost}
Unlike traditional ICP problems where source and reference point clouds belong to the same entity with variations possibly arising from partial observations, point clouds of gripper and object are likely to be quite distinct. The proposed algorithm approximates shape matching by maximizing the selected grasp quality metrics rather than achieving a perfect match.
These grasp quality metrics are selected to achieve two main objectives: minimize the matching error to estimate the grasp pose ($\mathcal{L}_{ct}$) and minimize the moment at contact points to ensure a stable grasp ($\mathcal{L}_{com}$):
 \begin{itemize}[leftmargin=*]
    \item $\mathcal{L}_{ct}$ minimizes the distance between the object's and the gripper's point clouds to estimate the grasp pose. Pairing points with zero distance can be considered as contact points of grasping. Thus, minimizing the least square distance can maximise the contact surface. For pairs of points $Pairs=\{s'_i,\hat{r}_i\}_{i=0}^{m}$ obtained from \eqref{eq:matching} using $\mathcal{S}$ and a mini-batch of size $m$ from $\mathcal{R}$, we optimize the following SGD-ICP point-to-point distance metric:
    \begin{equation}
    \begin{aligned}
        \mathcal{L}_{ct} = &\frac{1}{m}\sum_{ s'_i, \hat{r}_i \in \text{Pairs}}({\|s'_i-\hat{r}_i\|^2}).
    \label{eq:Lcontact}
    \end{aligned}
    \end{equation}

    \item $\mathcal{L}_{com}$ minimizes the distance between the gripper's tool centre point (TCP), which is defined as the center of the gripper's contact surface point cloud, and the object's centre of mass (CoM), defined as follows:
    \begin{equation}
     \mathcal{L}_{com} = \min\|(R \cdot TCP + t) \textendash CoM\|^2.
        \label{eq:lcom}
    \end{equation}
    
\end{itemize}

We use quaternions, unlike the original SGD-ICP~\cite{c2, c3} that uses Euler angle representation for rotations. Using the loss function $\mathcal{L}$, at iteration $k$ we have an average gradients of the translation components as
\begin{equation}
    \begin{aligned}
     \bar{g}(\theta_k^{1:3},\mathcal{S}_k) &= \frac{1}{m}\Bigl\{\sum_{ s'_i, \hat{r}_i \in \text{Pairs}}  {(s'_i-\hat{r}_i)}\frac{\partial t_k}{\partial \theta_k^{1:3}} \\
&+ ((R_kTCP + t_k) \textendash CoM)\frac{\partial t_k}{\partial \theta_k^{1:3}}\Bigl\},
\label{eq:grasp_trans_grads}
    \end{aligned}
\end{equation}
and for the rotation parameters, we have
\begin{equation}
    \begin{aligned}
       \bar{g}(\theta_k^{4:7},\mathcal{S}_k) &= \frac{1}{m}\Bigl\{\sum_{ s'_i, \hat{r}_i \in \text{Pairs}}  (s'_i-\hat{r}_i)\frac{\partial R_k}{\partial \theta_k^{4:7}}s_i \\
&+ ((R_kTCP + t_k) \textendash CoM)\frac{\partial R_k}{\partial \theta_k^{4:7}}TCP\Bigl\}.
\label{eq:grasp_rot_grads}
    \end{aligned}
\end{equation}
Partial derivatives of the rotation matrix  with respect to each component of the quaternion ($\frac{\partial R_k}{\partial \theta_k^{4:7}}$) are derived using the formulation presented in \cite{c341}.
\subsection{Collision Checking}
The ICP algorithm does not account for collision checking during the optimization process, which is crucial in grasping tasks. Simple ICP matching may cause the gripper to penetrate the object, resulting in physically infeasible grasp poses. Our objective is to efficiently move the gripper out of the object’s inner part in case of a collision while simultaneously running other AS-ICP samples that yield non-collision results in parallel for efficient grasp pose updates. We utilize the gripper's signed distance field (SDF)  which will return a positive distance value in equation  \eqref{eq:icp-objective} when there is a collision. Our method finds the closest matching points from $r_{col} \in\mathcal{C}$ to $\mathcal{S'}$, where $r_{col}$ are the points colliding with gripper. This approach contrasts with finding matches from $\mathcal{S'}$ to $\mathcal{R}$ as done in equation \eqref{eq:matching}. 
The loss function in equation \eqref{eq:total_weighted_loss} is replaced by the distance between $r_{col}$ and their nearest points on the gripper’s contact surface $s'$, as given below:

\begin{equation}
    \begin{aligned}
    \mathcal{L}_{col}& = \frac{1}{N_{col}}\sum_{r_{col} \in \mathcal{C} }\min_{s'\in \mathcal{S}}\|r_{col} \textendash s'\|^2, \\
    &s.t. \quad dist(\mathcal{C}, SDF(T(\mathcal{G}_i))) > 0,
        \label{eq:Lcol}
    \end{aligned}
\end{equation}
where $N_{col}$  is the number of colliding points in the object point cloud. In equation \eqref{eq:Lcol}, we do not account for $\mathcal{L}_{com}$ as the gripper needs to move away from the object’s center of mass (CoM). The gradients of this point-to-point cost function are obtained using equations \eqref{eq:trans_grad} and \eqref{eq:quat_grad}.

\SetKwComment{Comment}{-- }{}
\begin{small}
\begin{algorithm}[h!]
    \caption{Parallel Shape Matching with AS-ICP}\label{alg:sigsvgd}
    \SetAlgoLined
  	\DontPrintSemicolon
    \SetCommentSty{small}
    \KwIn{
    Point cloud of: 
    Gripper $\mathcal{S} = \{s_i\}_{i =1}^{N}$,
    Target Object $\mathcal{R} = \{ r_i\}_{i =1}^{M}$, 
    and Table $\mathcal{C}$;
    complete gripper $\mathcal{G}$,
    initial parameters ${\Theta}_0=\{\theta_0^j\}_{j=1}^{J}$,
    number of iteration for AS-ICP $k_{stein}$, and total iteration $k_{max}$
    }
    \KwOut{
        $\theta$ that minimizes the loss function 
    }

    $CoM \gets $ center of mass of target object\;
    $SDF \gets$ signed distance field of the gripper  \;
 
    \While{k $\leq k_{max}$}
{
      \For {each  $\theta_k^j\in \Theta_k$ in parallel }
   {
        $\mathcal{S}_k^j$  $\leftarrow $ Transform the source cloud with $\theta_k^j $\;
        $\mathcal{M}_k$ $\leftarrow $ Select a mini-batch 
        from $\mathcal{R}$\;
        $\bar{g}_k^j$ $\leftarrow$ Compute gradients \eqref{eq:grasp_trans_grads} and \eqref{eq:grasp_rot_grads}  \;
        \If{$dist(\mathcal{C}, SDF(T_{\theta^j_k}(\mathcal{G}))) > 0$}
        {
            $\bar{g}_{col}^j$ $\leftarrow$ Compute gradients using \eqref{eq:trans_grad} and \eqref{eq:quat_grad} \; 
            $\bar{g}_k^j$ = $\bar{g}_{col}^j$  $\leftarrow$ Replace $\bar{g}_k^j$  with $\bar{g}_{col}^j$ \;                                
       }
        \eIf{$k < k_{stein}$}
            {
            $ \hat{\phi}^*(\theta)$ $\leftarrow$ Compute using \eqref{eq:svgd-icp} \;
            $\theta_{k+1}^j$ $\leftarrow$ Update $\theta_{k}^j$ with \eqref{eq:svgd-update} \; 
            }   
            {
            $\theta_{k+1}^j$ $\leftarrow$ Update $\theta_{k}^j$ with \eqref{eq:sgdicp_update} \; 
            }   
   }
     k = k +1
}
        \Return $\theta = \operatorname*{argmin}_{\theta^j}\mathcal{L}$ 
\end{algorithm} 
\end{small}
\subsection{Algorithm}
In this work, we perform two optimization processes: AS-ICP and SGD-ICP. AS-ICP ensures that the poses are well-distributed around the object. After generating a distribution of potential grasping poses, we refine each Stein particle by updating it with SGD-ICP to determine the optimal grasp pose.

A summary of the Parallel Shape matching-based grasp method is provided in Algorithm \ref{alg:sigsvgd}. 
Given the object's point cloud $\mathcal{R}$ as a reference and the complete gripper's point cloud $\mathcal{G}$, we first compute the SDF of each preshape of the gripper and the target object's centre of mass.
We collect all preshapes' SDF as $\mathcal{SDF} = \{SDF_i\}^N_{i=1}$ by adding a constant offset $\epsilon$ between them, which can be considered as spreading them in space. All operations are performed in parallel for all initializations by combining the point clouds and parameters into multidimensional tensors $\mathcal{S}$, $\mathcal{M}$, $\mathcal{C}$ and $\mathcal{\theta}$. The source point cloud is transformed with the $J$ transformation parameters. Then, a mini-batch is sampled from the reference cloud. Next, corresponding closest points from the mini-batch are sought and stored in pairs for all points within each transformed source cloud, using Equation~\eqref{eq:matching}. This is followed by computing loss, gradients, and relative error difference. Instead of transforming the SDFs of the gripper, we perform an inverse transformation on the object point cloud for each initialization, denoted as $\mathcal{R'}=\{r_i'\}$. These point clouds are then divided into different groups according to the $SDF_i$ they are matching with. Then the same $\epsilon$ of the corresponding $SDF_i$ is added to $r'$. In case of collision, $\bar{g}_{col}$ is computed, replacing the current gradients. If the relative error difference indicates convergence at this stage, it is set to a non-converged state to move the gripper out of the object. Initially, AS-ICP is executed for $k_{\text{stein}}$ iterations. The resultant updated parameters $\theta$ are then used as the starting point for SGD-ICP. The parameter set $\theta$ undergoes further updates if it fails to converge. 

\section{Ablation Study}
\label{sec:ablation}
\begin{figure}[t]
    \centering
    \small
    \includegraphics[width=5cm]{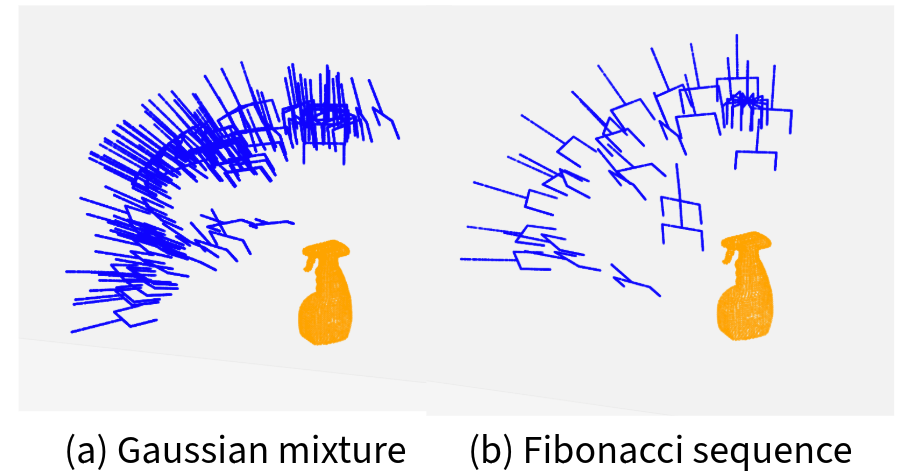}
    \caption{(a) Initializations sampled from a mixture of Gaussian to provide some prior knowledge of the object. (b) Initializations sampled from the Fibonacci sequence and projected onto a quarter of the sphere, facing the direction of the robot arm.}
    \label{fig:initialization}
\end{figure}
\begin{figure}[t]
    \centering
    \small
    \includegraphics[width=8.6cm]{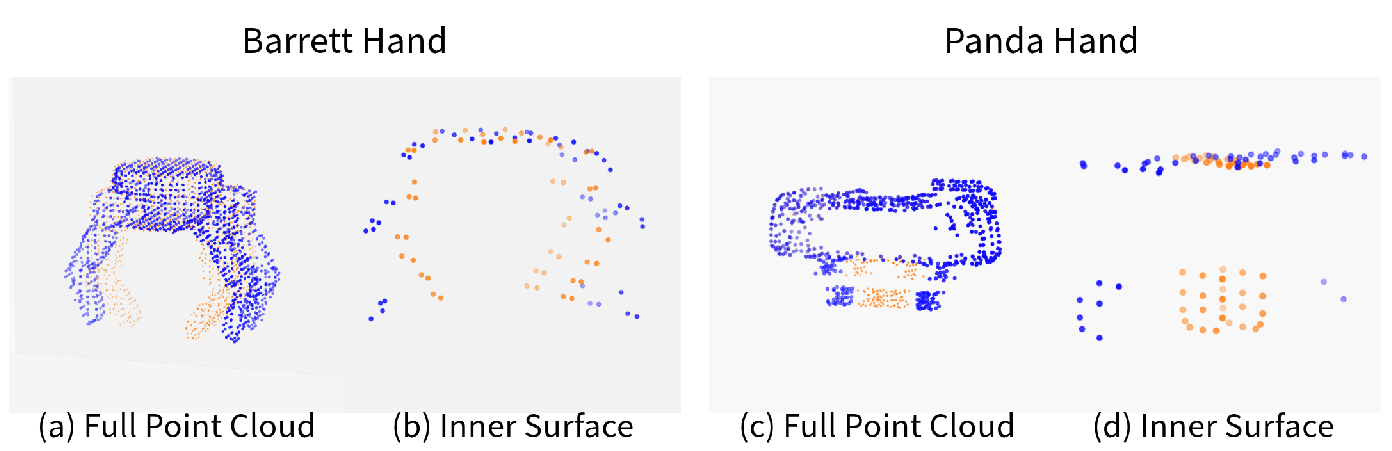}
    \caption{Preshapes used for simulation in this paper. On the left, we have two preshapes of Barrett Hand. On the right, we selected two preshapes from the ten used for Franka Hand. The full point cloud generates SDF, and the partial point cloud is used to optimize.}
    \label{fig:preshapes}
\end{figure}
\begin{figure}[t]
    \centering
    \small
    \includegraphics[width=0.85\linewidth]{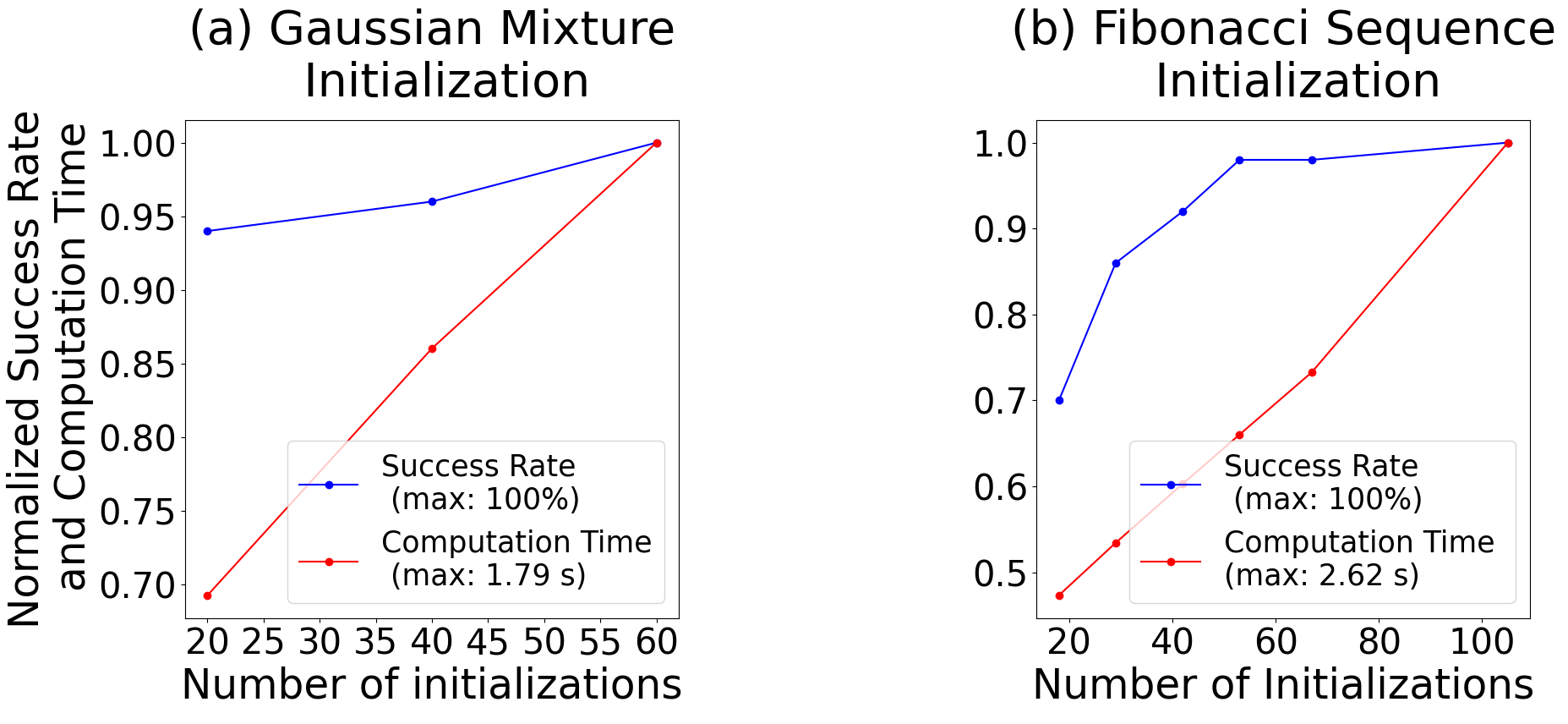}
    \caption{Plots of success rate and computation time with an increasing number of initializations for different sampling methods: (a) Gaussian and (b) Fibonacci. Sampling from a mixture of Gaussian leads to a higher success rate and faster computation time with fewer initializations as it provides some prior knowledge of objects.}
    \label{fig:n_particles}
\end{figure}

Initialization plays a pivotal role in ICP-based algorithms. Given its non-convex nature, ICP is susceptible to converge at local minima \cite{c26}. This aspect is particularly critical in our problem setup, where preshapes of the gripper vary significantly from the object geometry. Thus, precise initialization strategies are essential for achieving effective gripper-object matching.
In this work, AS-ICP effectively distributes initial poses around the object, increasing the likelihood of a successful grasp. Since we do not provide any prior knowledge of the objects, initializations are sampled from the Fibonacci sequence and projected onto a quarter of a sphere as shown in Fig. \ref{fig:initialization} (b). However, this introduces many unnecessary initializations and leads to longer computation time, as discussed in the previous section. A comparison between sampling from a mixture of Gaussians and Fibonacci sequences is provided in Fig. \ref{fig:n_particles}. It is worth noting that as the number of initializations increases, the success rate approaches 100\% at the cost of more computation time. By heuristically choosing four Gaussian means, our algorithm can achieve a higher success rate with fewer initializations, showing the possibility of learning some prior knowledge of the objects to reduce the dependency on initializations. 

In our algorithm, we match the gripper’s shape to the object's and do not optimize finger joints during the optimization process. Thus, choosing the right gripper preshapes becomes crucial. This paper uses two preshapes for the Barrett Hand and ten for Franka Hand, each corresponding to a set of fixed finger configurations, as shown in Fig. \ref{fig:preshapes}. Additional configurations are used for the Franka hand to ensure the stability of the generated grasps.

\section{Simulation and Real Experiment}

\subsection{Simulation Result}
\begin{figure}[t]
    \centering
    \small
    \includegraphics[width=7cm]{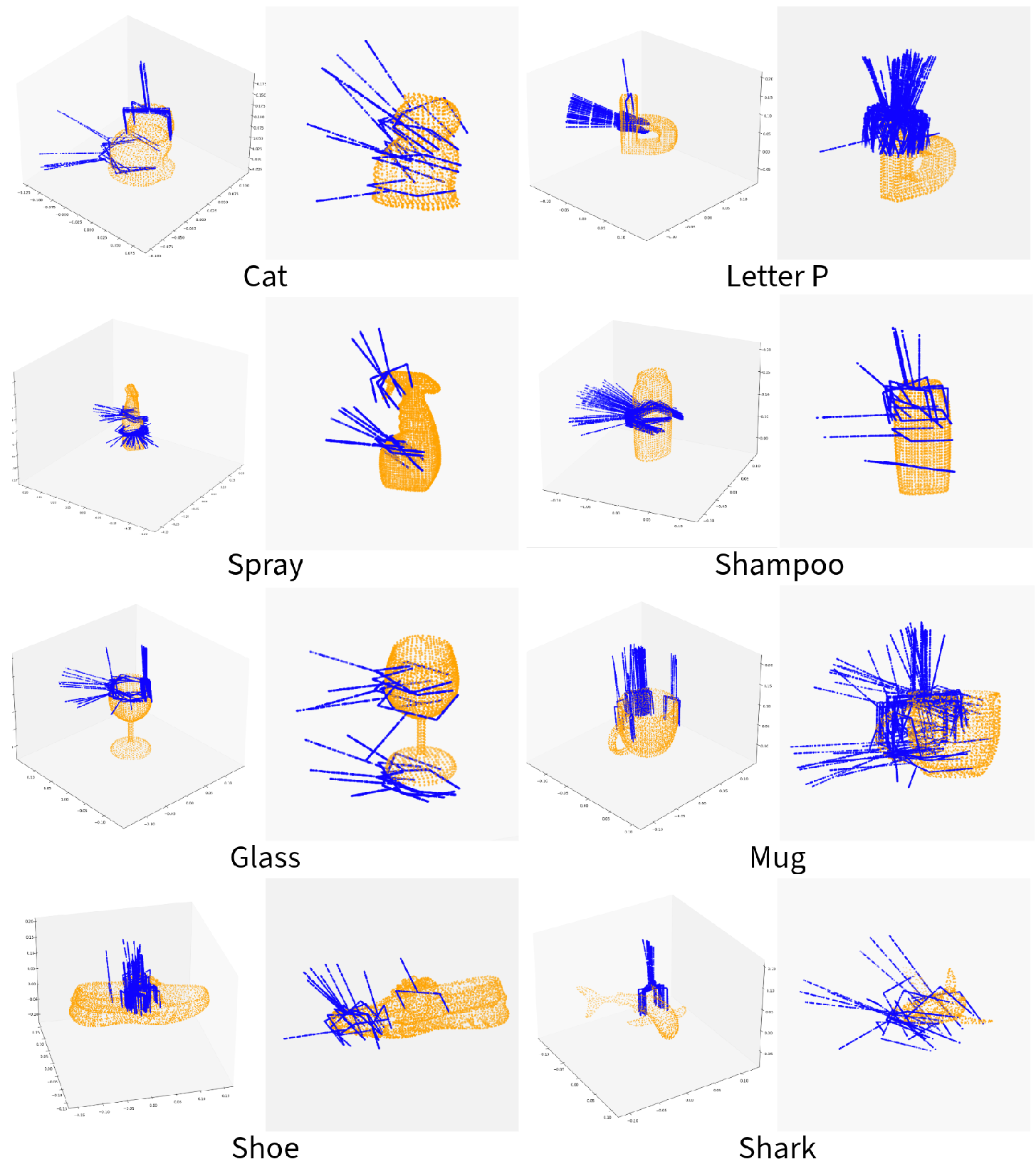}
    \caption{Best grasp poses with Franka Hand for 50 trials. On the left is the pose generated by our algorithm and on the right is the pose generated by AnyGrasp.}
    \label{fig:Panda}
\end{figure}

\begin{figure}[bt]
    \centering
    \small
    \includegraphics[width=7cm]{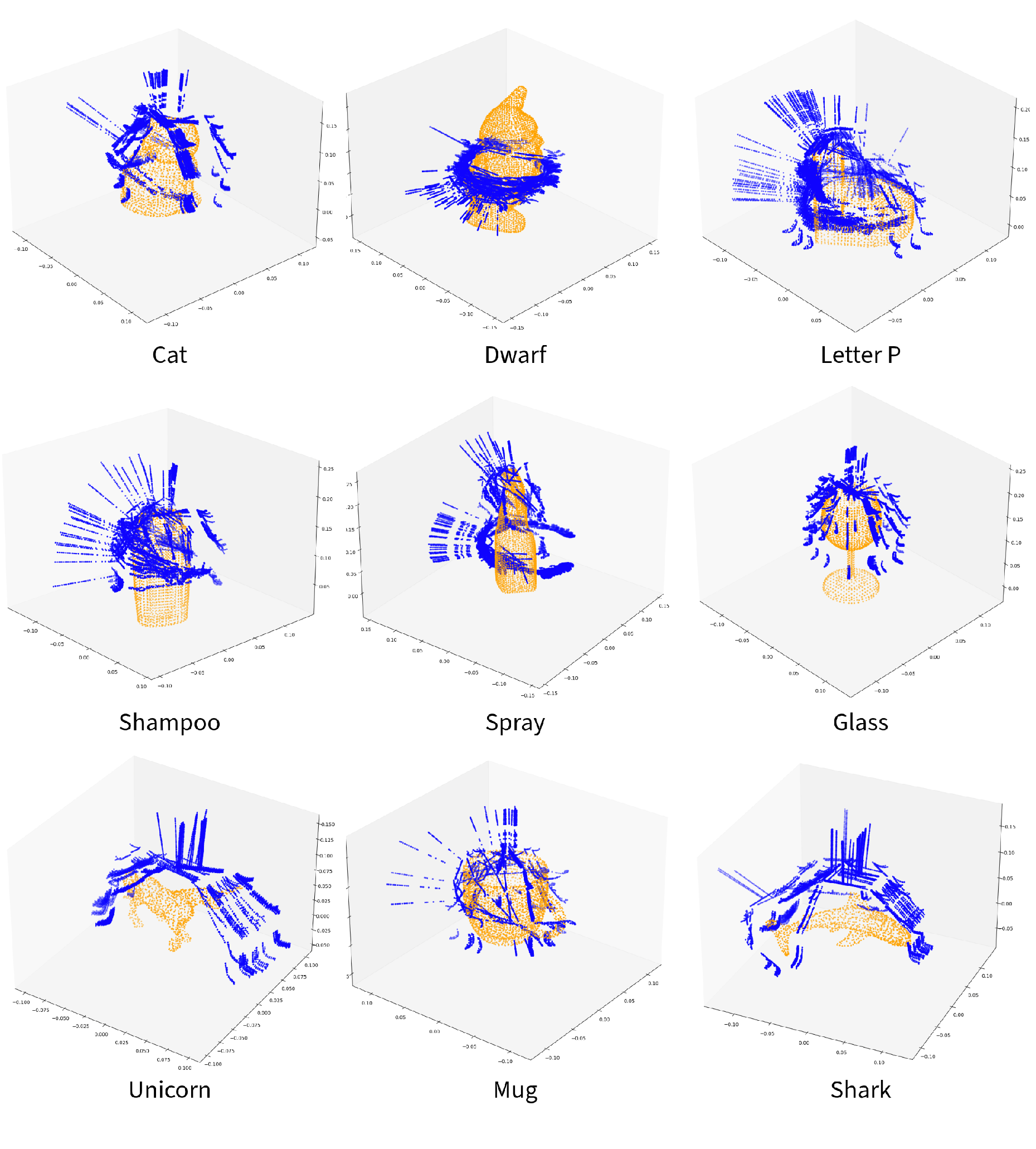}
    \caption{Best grasp poses with Barrett Hand for 50 trials.}
    \label{fig:BH}
\end{figure}

\begin{table}[t]
\small
\scriptsize
\centering
\begin{tabular}{|>{\centering\arraybackslash}m{5cm}|>{\centering\arraybackslash}m{1cm}|>{\centering\arraybackslash}m{1cm}|}
  \hline
   & Franka Hand & Barrett Hand \\
  \hline
  Number of initializations per preshape & \multicolumn{2}{c|}{$\approx \frac{100-6}{4}+6$} \\
  \hline
  Number of Preshapes  & 10 & 2 \\
  \hline
    Voxel Size (Gripper's point cloud $\mathcal{S}$)  & 0.005 & 0.025 \\
  \hline
  Number of Points (Object with table $\mathcal{C}$)  & \multicolumn{2}{c|}{1100 to 1900 points } \\
  \hline
    Voxel Size (Object's point cloud $\mathcal{M}$)  & \multicolumn{2}{c|}{0.005} \\
  \hline
    Mini-batch size  & \multicolumn{2}{c|}{$N * \frac{\min(k, 2k_{max}/3)}{2k_{max}/3}$} \\
  \hline
    Learning rate and cost weights for SGD & \multicolumn{2}{c|}{1} \\
  \hline
    Annealing Factor & \multicolumn{2}{c|}{$(\frac{\mod(k,k_{max}/5)}{k_{max}/5})^2$} \\
  \hline
    Convergence Criterion & \multicolumn{2}{c|}{0.02\%} \\
  \hline
    Number of iteration for SVGD & \multicolumn{2}{c|}{15} \\
  \hline
    Number of iteration for SGD & \multicolumn{2}{c|}{25} \\
  \hline
\end{tabular}
\caption{A list of parameters used for Alg. \ref{alg:sigsvgd}.}
\label{table:Parameters}
\end{table}

\begin{table}[t]
\begin{center}
\small
\scriptsize
\setlength{\tabcolsep}{2pt} 
\begin{tabular}{@{}lcc|cc|cc@{}}
\toprule
\multicolumn{1}{l}{\multirow{2}{*}{}} & \multicolumn{2}{c|}{AnyGrasp} & \multicolumn{2}{c|}{Ours} & \multicolumn{2}{c}{AnyGrasp+Ours} \\ \cmidrule(lr){2-3} \cmidrule(lr){4-5} \cmidrule(l){6-7}

\multicolumn{1}{c}{}  
& rate   & time (s)         & rate           & time (s) & rate    & time (s)    \\ \midrule
\multicolumn{1}{l|}{Cat}
& 46\%   & \textbf{0.0462}  & \textbf{98\%}  & 2.42     & 20\%    & 1.13\\
\multicolumn{1}{l|}{Detergent}
& 78\%   & \textbf{0.0473}  & \textbf{96\%}  & 2.57     & 88\%    & 1.08\\
\multicolumn{1}{l|}{Bottle}
& 82\%   & \textbf{0.0443}  & \textbf{96\%}  & 1.92     & 94\%    & 1.04\\
\multicolumn{1}{l|}{LetterP}
& 84\%   & \textbf{0.0515}  & \textbf{96\%}  & 2.63     & 86\%    & 0.93\\
\multicolumn{1}{l|}{Shampoo}  
& 28\%   & \textbf{0.0515}  & \textbf{94\%}  & 2.54     & 38\%    & 1.11\\
\multicolumn{1}{l|}{Spray} 
& 68\%   & \textbf{0.0497}  & \textbf{92\%}  & 2.45     & 80\%    & 1.02\\
\multicolumn{1}{l|}{Glass} 
& 68\%   & \textbf{0.0488}  & \textbf{96\%}  & 2.70     & 74\%    & 1.15\\
\multicolumn{1}{l|}{Mug}  
& 14\%   & \textbf{0.0508}  & \textbf{96\%}  & 2.32     & 78\%    & 1.14\\
\multicolumn{1}{l|}{Shark} 
& \textbf{20\%} & \textbf{0.0433} & 2\%      & 2.17     & 28\%    & 1.21\\
\multicolumn{1}{l|}{Shoe}   
& 6\%    & \textbf{0.0597}  & \textbf{54\%}  & 2.51     & 6\%     & 1.17\\ \midrule
\multicolumn{1}{l|}{\textit{\textbf{Average}}}
& 49.5\% & \textbf{0.0493}  & \textbf{82.0\%}& 2.42     & 59.2\%  & 1.10\\ \bottomrule
\end{tabular}
\end{center}
\caption{Simulation result comparison between AnyGrasp and our method. The simulation uses Franka Hand and is carried out in the Isaac Gym simulator.}
\label{table:simulation_Panda}
\end{table}

\begin{table}[t]
\begin{center}
\small
\scriptsize
\begin{tabular}{@{}lcc|cc@{}}
\toprule
\multicolumn{1}{l}{\multirow{2}{*}{}} & \multicolumn{2}{c|}{SplitPSO} & \multicolumn{2}{c}{Ours}        \\ \cmidrule(l){2-5} 
\multicolumn{1}{c}{}                        & rate             & time (s)   & rate           & time (s)       \\ \midrule
\multicolumn{1}{l|}{Cat}              & 90\%             & 8.10       & \textbf{100\%} & \textbf{0.996} \\
\multicolumn{1}{l|}{Detergent}        & 90\%             & 10.4       & \textbf{100\%} & \textbf{1.28}  \\
\multicolumn{1}{l|}{Dwarf}            & 50\%             & 9.55       & \textbf{100\%} & \textbf{1.56}  \\
\multicolumn{1}{l|}{LetterP}          & 80\%             & 12.7       & \textbf{100\%} & \textbf{1.55}  \\
\multicolumn{1}{l|}{Milk}             & \textbf{100\%}   & 11.1       & \textbf{100\%} & \textbf{1.43}  \\
\multicolumn{1}{l|}{Shampoo}          & 85\%             & 10.6       & \textbf{92\%}  & \textbf{1.20}  \\
\multicolumn{1}{l|}{Spray}            & 85\%             & 9.45       & \textbf{86\%}  & \textbf{1.40}  \\
\multicolumn{1}{l|}{Glass}            & 50\%             & 8.05       & \textbf{96\%}  & \textbf{1.24}  \\
\multicolumn{1}{l|}{Unicorn}          & 30\%             & 8.30       & \textbf{98\%}  & \textbf{1.43}  \\
\multicolumn{1}{l|}{Mug}              & 95\%             & 9.00       & \textbf{96\%}  & \textbf{0.972} \\
\multicolumn{1}{l|}{Shark}            & 0\%              & 25.4       & \textbf{14\%}  & \textbf{1.12}  \\ \midrule
\multicolumn{1}{l|}{\textit{\textbf{Average}}} & 68.6\%           & 11.2       & \textbf{89.3\%}         & \textbf{1.29}           \\ \bottomrule
\end{tabular}
\end{center}
\caption{Simulation result comparison between SplitPSO and our method. The simulation uses Barrett gripper and is carried out in Isaac Gym simulator.}
\label{table:simulation_BH}
\end{table}

Our study begins with a comprehensive examination of our proposed method's success rate and efficiency, employing two distinct grippers—-the Franka Hand and the Barrett Hand within the Isaac Gym simulator \cite{c21}. A list of parameters is provided in Table. \ref{table:Parameters}. The optimization is performed on a laptop with an RTX 2070 GPU. Initializations are sampled from a quarter of a sphere using the Fibonacci sequence. Six additional initializations directly above the object are manually added, as the sampled initializations alone do not provide sufficient coverage.  
We use Open3D \cite{c22} for both SDF generation and collision checking. After reaching the maximum iteration, the converged $\theta$ with the minimum cost is selected as the final grasp pose.

A successful grasp is defined as when the gripper can lift and hold the object for 5 seconds. Computation time excludes the time required to generate the SDF of the gripper. The gripper’s point cloud is sampled from the mesh for the Franka Hand. The gripper’s point cloud is acquired through simulation with a depth camera for the Barrett Hand. Objects are selected from the KIT object models database \cite{c23} and Google Scanned Objects \cite{c30}. The grasp poses generated are illustrated in Fig. \ref{fig:Panda} for Franka Hand and Fig. \ref{fig:BH} for Barrett Hand. A summary of the success rate and optimization time is provided in Table. \ref{table:simulation_Panda} and Table. \ref{table:simulation_BH}. The success rate and optimization time are the averages of 50 simulations for each object.

Table. \ref{table:simulation_Panda} provides a comparison with AnyGrasp \cite{c31}. The grasp poses for AnyGrasp were generated using the code from \cite{c32} with pre-trained weights. Instead of using a point cloud captured from a camera, the full point cloud of the object was utilized. Fifty grasps were generated with random object positions to compute the metrics. As a data-driven method, AnyGrasp has a shorter computation time than ours, whereas our method achieves an overall higher success rate.

Fig. \ref{fig:Panda}  presents a comparison of grasp poses generated by AnyGrasp (right) and our method (left). The primary factors leading to failures include slipping (Cat, Shampoo, Glass, and Mug) and unstable grasp poses (Spray, Shampoo, and Shoe). These issues are more pronounced with the poses generated by AnyGrasp. The lower success rate of AnyGrasp highlights its limitations when dealing with unseen objects and various working scenarios.

To evaluate the effectiveness of our algorithm as a post-processing module, we used 10 grasps generated by AnyGrasp as initializations for our method (Table. \ref{table:simulation_Panda}). Except for the Cat, the combined method achieved a higher success rate than AnyGrasp alone and had a lower computation time compared to our method alone. The failure with the Cat was due to slip, highlighting our method's dependence on good initializations, similar to other ICP-based methods. However, the improvement in success rate with shorter computation time is promising, suggesting that better initialization parameters could further enhance performance, as discussed in Sec. \ref{sec:ablation}.

Table. \ref{table:simulation_BH} provides a comparison for SplitPSO \cite{c6}, with grasp poses being generated from \cite{c33}.  Instead of using the proposed point cloud completion module, the object's full point cloud is used. Twenty grasps for each object are simulated. Our method outperforms SplitPSO in both success rate and computation time.

With limited configurations and a full gripper's inner surface point cloud, our algorithm mainly focuses on power grasp but not precision grasp. SplitPSO also faces this problem, as it uses the full point cloud of the gripper's inner surface. Thus, both methods fail in grasping the Shark, where a precision grasp with fingertips is more suitable. In contrast, AnyGrasp has a higher success rate, indicating a good precision grasp. Although the influence on a parallel gripper should be less significant, using the palm point cloud still affects our success rate with Franka Hand. This can be seen from the grasp poses generated for the Shark, where our method fails due to contact with the Shark’s fin. A potential solution is to sample point clouds from the finger surface with some distribution, as proposed by \cite{c9}. However, this object-dependent parameter requires data-driven methods to acquire the best preshapes for different objects.

\subsection{Real Experiment}
\begin{figure}[t]
    \centering
    \small
    \includegraphics[width=7cm]{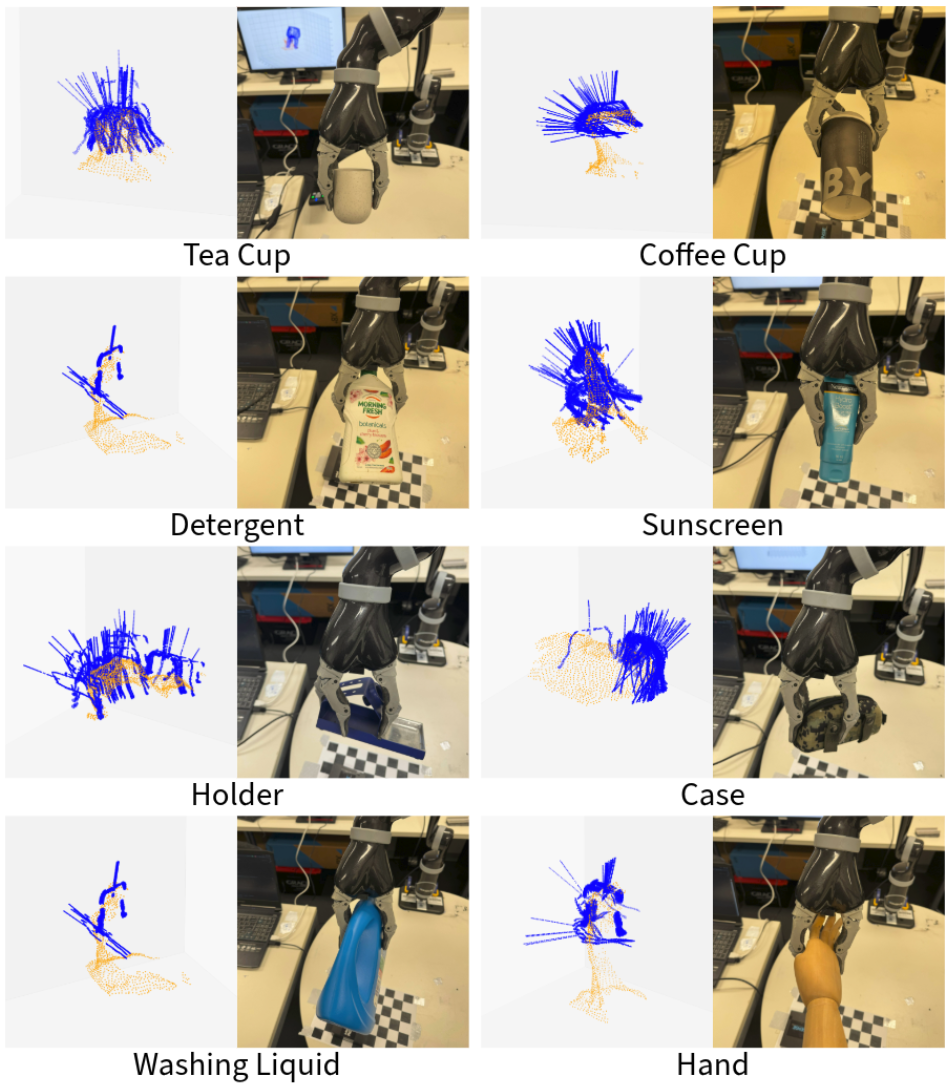}
    \setlength{\belowcaptionskip}{-12pt}
    \caption{Picture on the left illustrates fifty grasp poses generated with the same point cloud. The photo on the right is the actual grasp for one of the five grasps in Table. \ref{table:real}. Our algorithm can grasp objects with large occlusions and noisy point clouds.}
    \label{fig:Real}
\end{figure}

\begin{figure}[t]
    \centering
    \small
    \includegraphics[width=6cm]{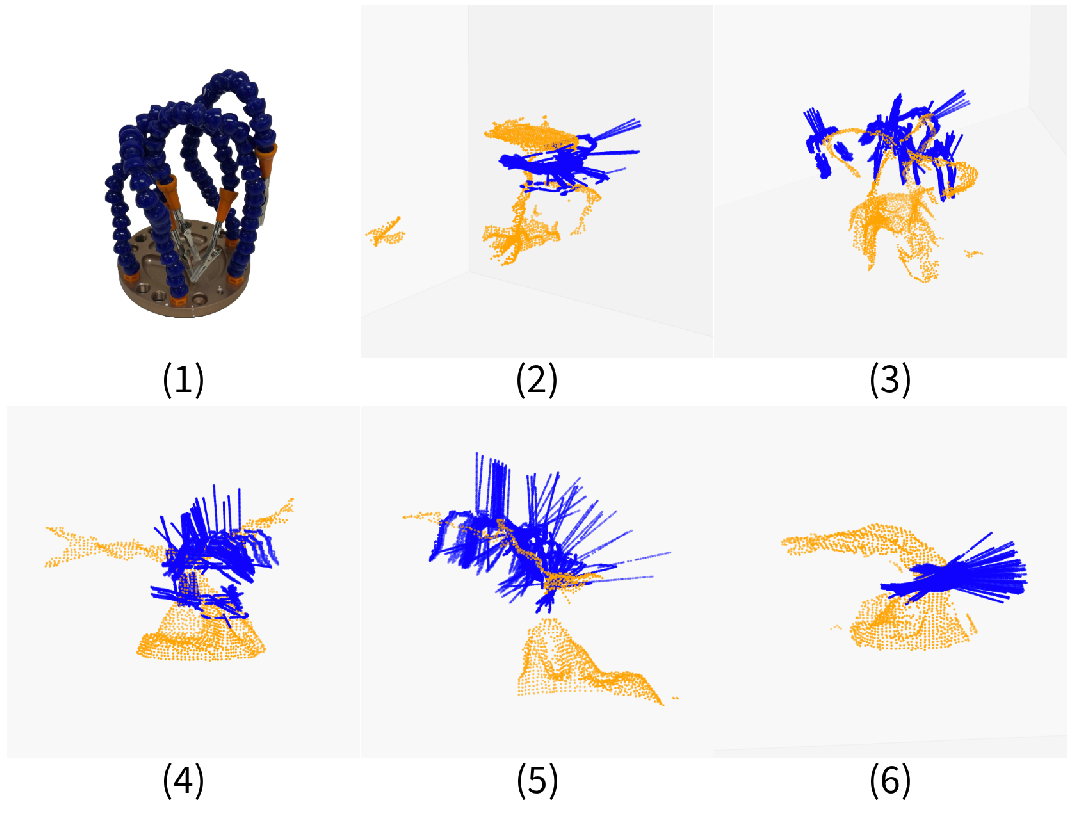}
    \setlength{\belowcaptionskip}{-12pt}
    \caption{(1) Hand Helping Tool (2)-(6) Grasp poses for Hand Helping Tool with five different shapes.}
    \label{fig:real_abl}
\end{figure}

\begin{table}[t]
\begin{center}
\small
\scriptsize
\begin{tabular}{lcc}
\toprule
                 & Success rate & Time (s) \\ \midrule
Tea Cup          & 100\%          & 1.01    \\
Coffee Cup       & 100\%          & 0.949    \\
Detergent        & 80\%          & 0.616    \\
Brush            & 100\%          & 0.989    \\
Sunscreen        & 100\%          & 0.842    \\
Holder           & 60\%          & 1.06    \\
Case             & 80\%          & 1.06     \\
Washing Liquid   & 100\%          & 0.812    \\
Toy              & 80\%          & 1.11     \\
Hand             & 80\%          & 0.792     \\
Helping Hand Tool& 80\%          & 0.952    \\ \midrule
\textit{Average} & 87.3\%       & 0.926    \\ \bottomrule
\end{tabular}
\end{center}
\caption{Number of successes for five grasps, where the input for each grasp corresponds to a different orientation of the object.}
\label{table:real}
\vspace{-5mm}
\end{table}

We conducted real experiments using the Jaco 2 Arm equipped with the KG3 gripper to test our approach with noise and partial point clouds. We captured objects' point clouds using a wrist-mounted RealSense D405 Camera. Unlike in simulation, where the algorithm has access to the entire point cloud, we used a single viewpoint partial point cloud for the real experiment.

The experimental results, shown in Fig. \ref{fig:Real}, demonstrate the performance of our approach on eight objects. Five grasps were executed for each object using the same parameters and initializations as in the simulation. Each trial involved placing the object in a different orientation, resulting in varied occlusions. A waypoint was set 15 cm away from the grasp pose to prevent collisions during motion. After grasping, the gripper returned to the location indicated in Fig. \ref{fig:Real}. On average, we could grasp 4.37 out of the 5 grasps, with an average computation time of 0.926 seconds, as summarized in Table \ref{table:real}. 

Notably, challenges arise in scenarios with large occlusions in the object point cloud, contributing to most fail cases with objects such as Detergent and Washing Liquid. To further challenge our algorithm, we tested it with multiple shapes of the Hand Helping Tool, as shown in Fig. \ref{fig:real_abl}.  As the point cloud does not provide physical properties of the object, some grasp poses will fail on the flexible regions of the object, as indicated by the failure of the Hand and Hand Helping Tool, where the former failed due to flexible fingers and the latter failed due to a false joint, as shown in Fig. \ref{fig:real_abl} (3). In addition, objects requiring precision grasp also pose challenges in real experiments. For example, our algorithm can grasp the Toy while standing but fails when lying on the table.

\section{CONCLUSIONS}
We have introduced a rigid shape-matching grasp optimization method that utilizes the parallel AS-ICP algorithm. Demonstrating its versatility across various gripper types, the approach exhibits remarkable robustness against noisy and partial point clouds, achieving an average computation time of 0.926 seconds and an average success rate of 87.3\% for the KG3 gripper in real experiments surpassing other point cloud-based analytic approaches.
Our results underscore the potential of parallel optimization methods to enhance grasp quality and reduce computation time in robotic manipulation tasks. Furthermore, we also observed that optimizing gripper joint angles is not compulsory to ensure a successful grasp. 

Future research will integrate data-driven approaches for gripper preshape selection and initial pose generation, further reducing computation time while enhancing grasp quality.


\addtolength{\textheight}{-9cm}


\bibliographystyle{ieeetr}
\bibliography{references}

\end{document}